\documentclass[letterpaper]{article} 
\usepackage{aaai2026}
\usepackage{times}  
\usepackage{helvet}  
\usepackage{courier}  
\usepackage[hyphens]{url}  
\usepackage{graphicx} 
\usepackage{natbib}  
\usepackage{caption} 
\usepackage{xcolor}
\usepackage{adjustbox}
\usepackage{multirow}
\usepackage{algorithm}
\usepackage{algorithmic}
\usepackage{amsmath} 
\usepackage{amssymb}
\usepackage{array}
\newcolumntype{S}{>{\small}c}
\usepackage{newfloat}
\usepackage{listings}
\DeclareCaptionStyle{ruled}{labelfont=normalfont,labelsep=colon,strut=off} 
\floatstyle{ruled}
\newfloat{listing}{tb}{lst}{}
\floatname{listing}{Listing}
\title{Unleashing Hierarchical Reasoning: An LLM-Driven Framework for Training-Free Referring Video Object Segmentation}
\author{
    Bingrui Zhao\textsuperscript{\rm 1},
    Lin Yuanbo Wu \textsuperscript{\rm 2} \thanks{Corresponding author.},
Xiangtian Fan \textsuperscript{\rm 1},
Deyin Liu \textsuperscript{\rm 3} ,
Lu Zhang \textsuperscript{\rm 2}, 
Ruyi He \textsuperscript{\rm 1}, 
Jialie Shen \textsuperscript{\rm 4},
    Ximing Li \textsuperscript{\rm 1}
}
\affiliations{
    \textsuperscript{\rm 1}Jilin University, China\\

    \textsuperscript{\rm 2} Swansea University, United Kingdom\\
    \textsuperscript{\rm 3} Anhui University, China\\
   \textsuperscript{\rm 4}  City St George's, University of London, United Kingdom

}

\usepackage{bibentry}

\begin{document}

\maketitle

\begin{abstract}
Referring Video Object Segmentation (RVOS) aims to segment an object of interest throughout a video based on a language description. The prominent challenge lies in aligning static text with dynamic visual content, particularly when objects exhibiting similar appearances with inconsistent  motion and poses. However, current methods often rely on a holistic visual-language fusion that struggles with complex, compositional descriptions. In this paper, we propose \textbf{PARSE-VOS}, a novel, training-free framework powered by Large Language Models (LLMs), for a hierarchical, coarse-to-fine reasoning across text and video domains. Our approach begins by parsing the natural language query into structured semantic commands. Next, we introduce a spatio-temporal grounding module that generates all candidate  trajectories for all potential target objects, guided by the parsed semantics. Finally, a hierarchical identification module select the correct target through a two-stage reasoning process: it first performs coarse-grained motion reasoning with an LLM to narrow down candidates; if ambiguity remains, a fine-grained pose verification stage is conditionally triggered to disambiguate. The final output is an accurate segmentation mask for the target object. \textbf{PARSE-VOS} achieved state-of-the-art performance on three major benchmarks: Ref-YouTube-VOS, Ref-DAVIS17, and MeViS.
\end{abstract}


\section{Introduction}
\begin{figure}[h]
    \centering
    \includegraphics[width=0.95\columnwidth]{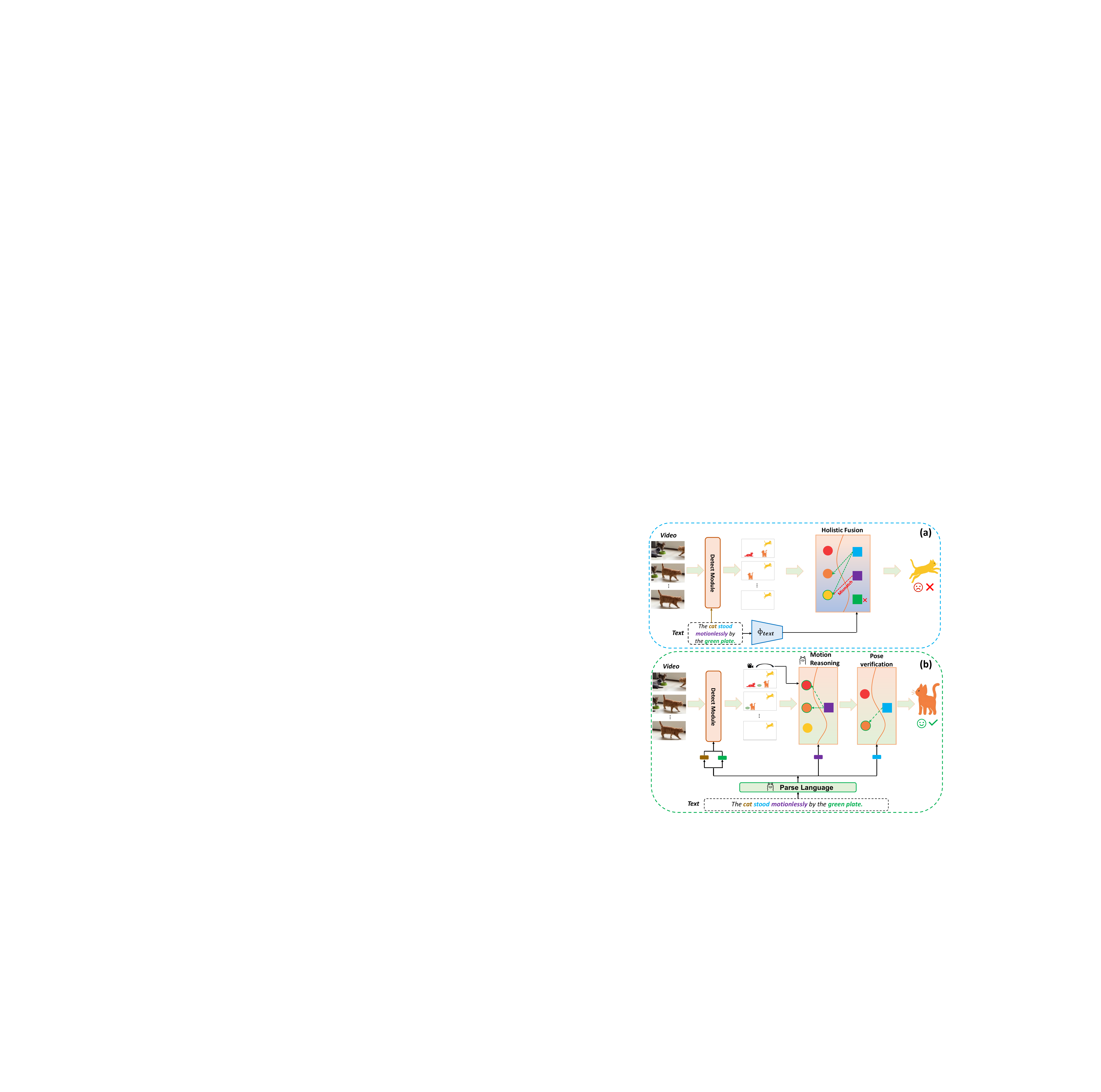}  
    \caption{Comparison between our hierarchical reasoning framework (b) and traditional holistic alignment method (a). Unlike holistic alignment, our method can extract rich video contextual features by first screening candidate objects through coarse-grained motion inference, and then using fine-grained pose validation to address ambiguity. This structured approach can effectively handle complex scenarios that traditional methods find difficult to handle.}
    \label{fig:introduct_figure}
\end{figure}

Referring Video Object Segmentation (RVOS) is a fundamental visual-language task that segments objects in a video based on natural language descriptions, and thus enables a more intuitive, human-centric interaction. This approach shows great potential in applications like video editing \cite{Bar-Tal2022Text2live,Liu-PRAI-2024}, human-computer interaction \cite{Li2023Robust}, autonomous driving assistance systems \cite{Gao2025LangCoop,Gao2025Automated}, and robot perception \cite{Qi2020Reverie,Qin2023Unified}.

Mainstream RVOS methods fall into two primary paradigms: end-to-end holistic fusion based approaches and ``detect-then-filter" approaches. Traditional end-to-end methods \cite{seo2020urvos,khoreva2018video,wu2022language,ZhongICME-2025,CTVIS-2023}, particularly those based on Transformer\cite{transfomer}, employ powerful attention mechanisms in an encoder-decoder structure to perform a ``holistic" fusion, attempting to directly align language semantics with visual features. However, when faced with complex or compositional language (such as descriptions involving negations or subtle attribute differences) , this holistic fusion strategy struggles to robustly map fine-grained linguistic structures to dynamic visual content, leading to a ``semantic gap". To address this, the ``detect-then-filter" paradigm emerged. These methods\cite{he2024decoupling,Kim2025SOLA} first segment all potential object candidates in a video and then use a separate module to select the target based on the language description.
While this divide-and-conquer strategy has alleviated the challenges of direct alignment to some extent, its decoupling of perception from reasoning introduces a new challenge: it tends to evaluate each candidate object in isolation and thus ignores rich surrounding context. A language description often contains not only the target's attributes but also its complex relationships with other entities and the overall scene dynamics. For instance, in Figure \ref{fig:introduct_figure}, the language instruction comes as ``The cat stood motionlessly by the green plate". To correctly interpret this prompt, the model must not only identify the ``cat" but also understand its spatial proximity to the contextual entity, i.e., the ``green plate". When there is camera panning, the description ``motionlessly" becomes highly ambiguous. In this context, models that analyze isolated trajectories without accounting for camera motion cannot accurately determine which targets are truly stationary relative to the scene. This limitation in reasoning about complex contextual dynamics highlights a key weakness of conventional ``detect-then-filter” methods.

To address the aforementioned limitations, we propose PARSE-VOS, a novel, training-free framework that recasts the task of Referring Video Object Segmentation as a hierarchical, coarse-to-fine reasoning process. Unlike approaches that attempt direct end-to-end fusion, our framework is based on ``parse-and-reasoning". First, we leverage a Large Language Model to parse the complex natural language instruction into structured, machine-readable semantic commands. With these interpretable commands, the model localizes and tracks all relevant candidate objects in the video, generating their spatio-temporal trajectories. Then we develop a hierarchical reasoning and identification pipeline, which can effectively tackle referring expressions that mix coarse motion cues with fine-grained visual attributes. This hierarchical process first employs coarse-grained motion reasoning to eliminate easy distractors, followed by a conditional fine-grained pose verification to resolve hard ambiguity. Inspired by LLM inference based motion trajectory \cite{guiji1,guiji2,guiji3}, we employ the LLM as a zero-shot spatio-temporal reasoner and introduce contextual priors, such as camera motion and occlusion relationships, to comprehend the scene's dynamics and spatial hierarchy. The LLM first performs coarse-grained motion reasoning to swiftly eliminate a large number of irrelevant candidates. Only if ambiguity persists, a fine-grained pose verification stage is conditionally activated to resolve ambiguity. This fully training-free architecture not only allows us to handle complex scenes and language descriptions but also circumvents the reliance on large-scale training data and overfitting problems.

\begin{itemize}
    \item We propose PARSE-VOS, a novel, training-free, hierarchical reasoning framework  powered by a LLM for the task of RVOS.
    
    \item We introduce an LLM-based reasoning mechanism enhanced by contextual priors like camera motion and occlusion,empowering the model with sophisticated spatio-temporal reasoning to disambiguate targets.
    
    \item Our method achieves state-of-the-art performance on three major RVOS benchmarks (Ref-YouTube-VOS, Ref-DAVIS17, and MeViS) , demonstrating the effectiveness and superiority of our proposed framework, particularly in complex and highly ambiguous scenarios.
\end{itemize}

\section{Related Work}

\subsection{Referring Video Object Segmentation} Referring Video Object Segmentation (RVOS) is a fundamental visual-language task that aims to segment an object of interest in a video based on a natural language description. The task was introduced with the proposal of the A2D-Sentences dataset \cite{Gavrilyuk2017Actor} and has since flourished, propelled by the establishment of influential benchmarks like Ref-DAVIS \cite{khoreva2018video} and the large-scale Ref-YouTube-VOS \cite{seo2020urvos}. This progress has, in turn, spurred the creation of new datasets targeting more complex scenarios, such as MeViS \cite{ding2023mevis}.

End-to-end methods, such as ReferFormer \cite{wu2022language} and MTTR \cite{botach2022end}, leverage Transformer-based architecture to achieve a holistic alignment between language and vision via attention mechanism. While this strategy streamlines the pipeline, the holistic fusion is lacking the ability to map fine-grained linguistic structures, leading to a ``semantic gap" in ambiguous scenarios. Another dominant paradigm is the ``detect-then-filter” approach, exemplified by methods such as ReferDINO \cite{liang2025referdino} and SOLA \cite{Kim2025SOLA}, which leverage powerful visual foundation models like GroundingDINO \cite{liu2023grounding} and SAM2 \cite{ravi2024sam2segmentimages} to achieve impressive performance. However, this approach often neglects the global temporal context of the video, and its reliance on large-scale training data for filtering introduces risks of overfitting and limited generalization.

\subsection{LLMs in Segmentation and Grounding} The proliferation of Large Language Models (LLMs) has catalyzed their application in vision tasks, often in the form of Multi-modal Large Language Models (MLLMs). For instance, methods like Video-Lisa \cite{bai2024one} utilize specialized tokens to enable language-driven segmentation and tracking, VISA \cite{yan2024visa} Use a Text Guided Frame Sampler to filter out the most relevant keyframes from videos based on complex text instructions, and then input these frames along with the text into a MLLM to generate segmentation instructions that include specific ones, and ViLLa\cite{zheng2024villa} prepares refined visual and textual features for large language models through its designed extractors and synthesizers, and ultimately decodes them into precise video masks through a layered time synchronizer. While these MLLM-based approaches achieve remarkable performance, their substantial computational cost presents a barrier to practical application. In sharp contrast, we uniquely employ a pure LLM as a dedicated spatio-temporal reasoner that processes structured textual information derived from the video, thereby striking an effective balance between high performance and practical accessibility.

\begin{figure*}[t]
    \centering
    \includegraphics[width=1.0\textwidth]{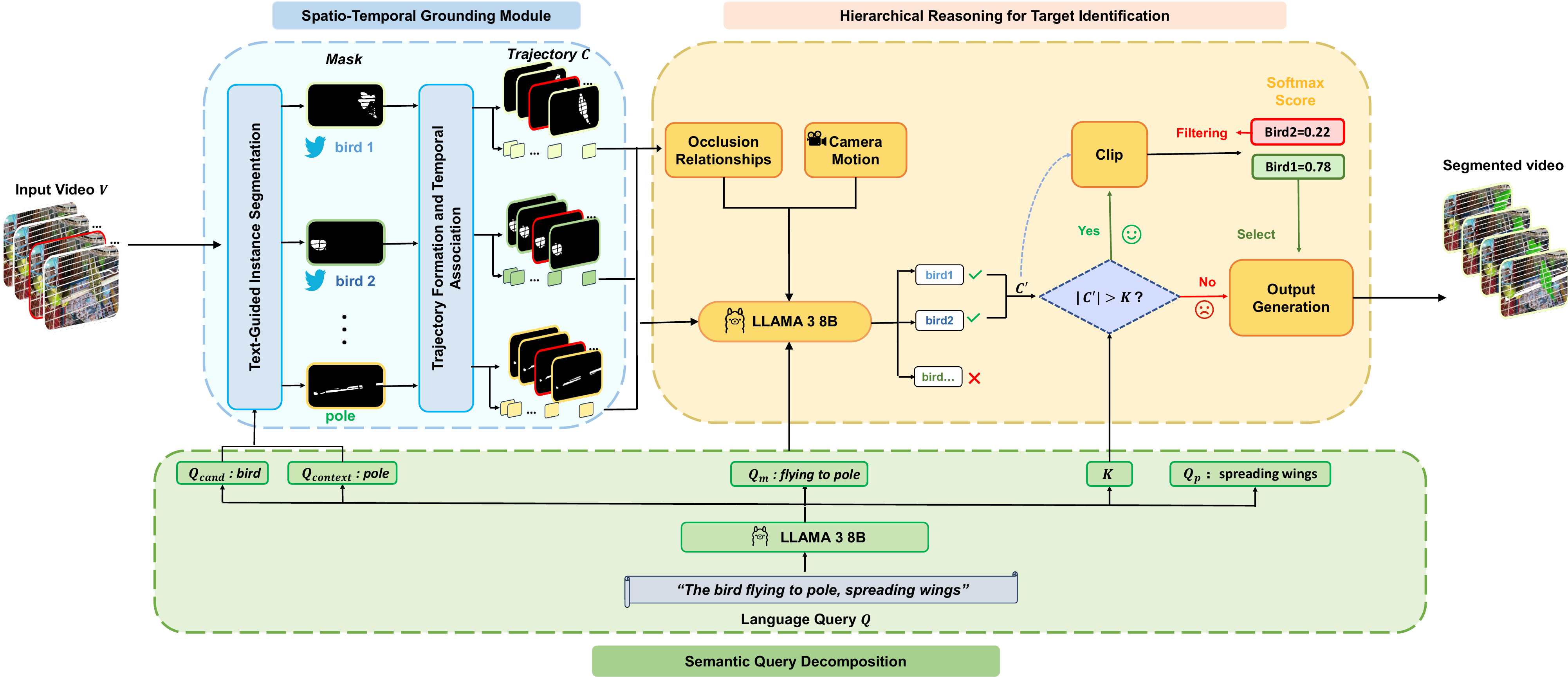}
    \caption{Overview of the proposed hierarchical, coarse-to-fine framework (PARSE-VOS). The pipeline primarily consists of \textbf{(M1) Semantic Query Decomposition:} We use the LLM to parse the natural language query $Q$ into structured commands. \textbf{(M2) Spatio-Temporal Candidate Grounding:} Potential target objects are detected and tracked throughout the video to form candidate trajectories $\mathcal{C}$. \textbf{(M3) Hierarchical Target Identification:} A two-stage reasoning process, which involves coarse-grained motion reasoning and conditional pose verification to identify each target trajectory $c^*$ from $\mathcal{C}$ and generate the output mask sequence $M$.}
    \label{fig:my_wide_figure}
\end{figure*}

\section{Method}\label{sec:method}
The Referring Video Object Segmentation (RVOS)  aims to segment the target object(s) queried by a natural language description $Q$ within a video sequence $V$. The core challenge of this task lies in accurately aligning the static language description with dynamically changing object instances in both temporal and spacial landscape. Given a video sequence $V$ composed of $T$ frames, i.e., $V = \{I_t\}_{t=1}^T$, where each frame may contain multiple object instances $O = \{o_i\}_{i=1}^N$, the goal is to learn a mapping function $\Phi$ which takes the video $V$ and the query $Q$ as input to generate a binary mask $M_t$ for each frame, and thus yielding the mask sequence $M = \{M_t\}_{t=1}^T$.

In this paper, we propose a novel framework named PARSE-VOS, which formulate the RVOS task as a \textbf{hierarchical, coarse-to-fine} reasoning process. As illustrated in Figure \ref{fig:my_wide_figure}, the proposed PARSE-VOS locates the targets via cascaded filtering and reasoning, which progressively involves three following on-site modules.

\subsubsection{M1: Semantic Query Decomposition}

This module takes the original natural language query $Q$ and utilizes a large language model (Llama 3 8B) to parse it into a set of structured commands including candidate entities ($Q_{\text{cand}}$), contextual entities($Q_{\text{context}}$), motion descriptors ($Q_m$), posture/attribute descriptors ($Q_p$), and the expected target cardinality $K$. This structured output provides clear and disentangled guidance for the subsequent visual perception and reasoning modules.

\subsubsection{M2: Spatio-Temporal Candidate Grounding}

Guided by the parsed entity queries ($Q_{\text{cand}}$, $Q_{\text{context}}$), this module is responsible for localizing all potentially relevant objects in the video $V$. We first perform text-guided instance segmentation on sparse keyframes using an open-vocabulary detector (GroundingDINO) and a segmentation model (SAM2). These segmented instances are then temporally associated to form complete spatio-temporal trajectories for each potential target in the scene. The output of this stage is a set of candidate trajectories $\mathcal{C}$.

\subsubsection{M3: Hierarchical Target Identification}
This module identifies the final target from the candidate trajectories $\mathcal{C}$ through a two-stage, coarse-to-fine reasoning process. First, a \textbf{coarse-grained motion reasoning} stage leverages the motion command $Q_m$ to rapidly filter candidates based on trajectory consistency. Then, if ambiguity persists, a \textbf{fine-grained pose verification} stage is conditionally activated, using the posture command $Q_p$ to perform detailed visual-semantic alignment. This process pinpoints the final target trajectory $c^*$ and generates the output mask sequence $\mathcal{M}$.

\subsection{Semantic Query Decomposition}
To decompose the unstructured natural language query Q into structured commands for our downstream modules, we employ Llama 3 8B model \cite{meta2024llama3}, for its potent instruction-following and contextual understanding capabilities, as a zero-shot semantic parser, decomposing any given query Q into five distinct semantic components via a carefully designed prompt.These include candidate entities and their descriptions, $(Q_{\text{cand}})$, contextual entities, $(Q_{\text{context}})$, motion descriptors, $Q_m$, posture/attribute descriptors, $Q_p$, and the target cardinality $K$. This structured output provides precise and disentangled guidance for the subsequent modules of our framework.

\subsection{Spatio-Temporal Grounding Module}
\begin{figure*}[t]
    \centering
    \includegraphics[width=0.95\textwidth]{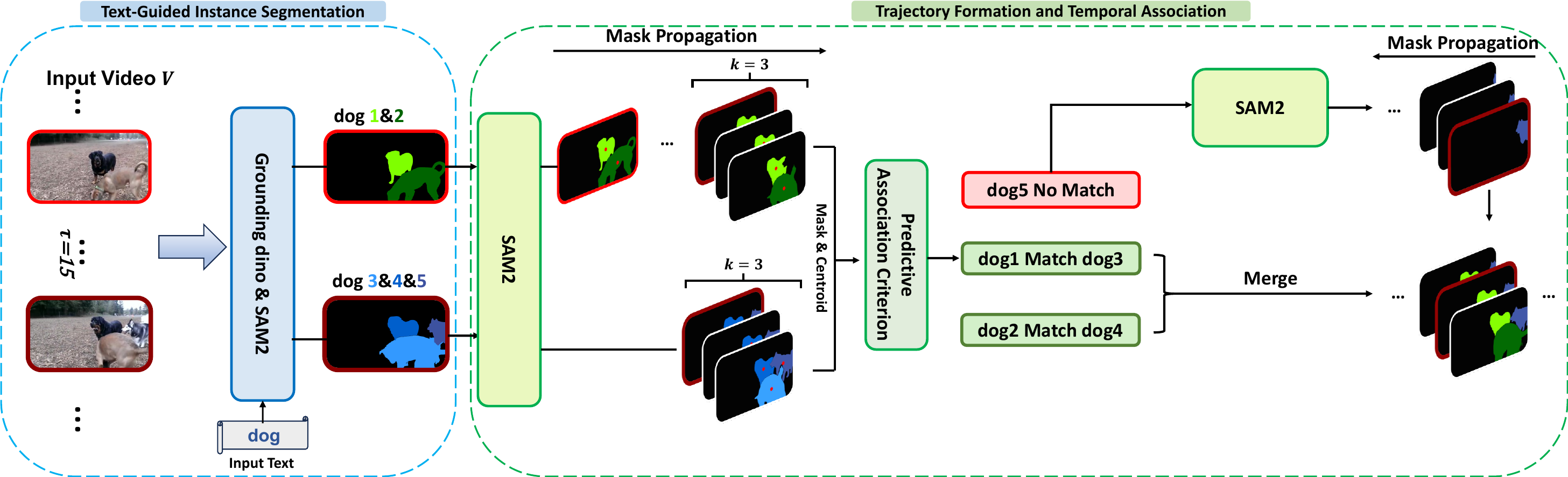} 
    \caption{
        An overview of our Spatio-Temporal Grounding Module. (Left) Text-Guided Instance Segmentation: We generate instance masks on sparse keyframes using a text-guided detector. (Right) Trajectory Formation and Temporal Association: Subsequently, a Predictive Association Criterion matches new detections with legacy tracks by comparing their propagated masks and centroids to form continuous trajectories.
    }
    \label{fig:grounding_module} 
\end{figure*}

The Spatio-Temporal Grounding module serves as the perceptual foundation of our framework. Its input is the raw video $V$ and the entity queries decomposed from the previous module (i.e., candidate targets $Q_{cand}$ and contextual objects $Q_{context}$). Its core task is to locate and continuously track all query-related objects through a clear two-step process. This process first performs \textbf{text-guided instance segmentation} in the spatial dimension on sparse keyframes, and then performs \textbf{trajectory formation and temporal association} to form continuous tracks. Ultimately, this module outputs a structured set of spatio-temporal trajectories $\mathcal{C}$, providing a reliable and dynamic scene representation for the subsequent reasoning module, as illustrated in Figure~\ref{fig:grounding_module}.

\subsubsection{Text-Guided Instance Segmentation}
To balance efficiency and accuracy, we perform instance recognition on keyframes sampled at a regular interval of $\tau$ frames ($\tau=15$ in our experiments), For each keyframe, we employ GroundingDINO \cite{liu2023grounding} as our open-vocabulary detector. The set of all entity nouns, $Q_{\text{cand}} \cup Q_{\text{context}}$,  is provided as the textual prompt to generate bounding boxes for all relevant objects in the scene. To achieve pixel-level precision, these bounding boxes are then fed as visual prompts into the SAM2\cite{ravi2024sam2segmentimages}, which outputs a high-fidelity binary mask for each detected instance.

\subsubsection{Trajectory Formation and Temporal Association}

This module aims to connect the static instance masks, obtained from the previous module on sparse keyframes, into temporally coherent trajectories. The lifecycle of each trajectory is managed within this module through an iterative process: a trajectory is initialized upon an object's first appearance, or extended through successful association with new instances in subsequent frames. In specific, this association is formulated by a proactive matching strategy where the SAM2 model is employed to propagate a legacy track from keyframe $t$ and a new instance from next keyframe $t+\tau$ forward, generating two short future paths. Subsequently, we quantify the similarity of these two predictive paths using two key metrics: average Intersection over Union (IoU) and centroid distance. Herein, the temporal association is governed by the proposed \textbf{predictive association criterion}, where a match is confirmed only when the average IoU over a short window is above a threshold $\theta_{iou}$ (e.g., 0.6) and the average centroid distance is below a threshold $\theta_{dist}$ (e.g., 50 pixels). If a match is successful, the corresponding legacy track is extended. Conversely, if a new instance fails to match any legacy track, a new trajectory is initialized for it, and its completeness is ensured via using SAM2 to retroactive propagation.

\subsection{Hierarchical Reasoning for Target Identification}
The hierarchical reasoning module is responsible for synthesizing the structured information provided by the upstream modules, i.e., the candidate trajectories $\mathcal{C}$ and the parsed commands ($Q_m, Q_p, K$), to precisely identify the final target. To better understand scene dynamics and spatial relationships during the reasoning process, this module internally generates and utilize contextual information such as a camera motion model and object depth relationships. To balance both efficiency and accuracy, the module operates on a coarse-to-fine principle realized through a two-stage cascade: first, a coarse-grained motion reasoner performs an initial, broad filtering; and then, a fine-grained pose verifier is conditionally activated to resolve any remaining ambiguities.

\subsubsection{Coarse-Grained Motion Reasoning}
The objective of this stage is to leverage an LLM to filter the full set of candidate trajectories $C$ down to a smaller subset $C'$ that is semantically consistent with the motion query $Q_m$. To prepare the motion data for this task, we begin with \textbf{Trajectory Textual Serialization}. This process encodes the apparent movement of each candidate trajectory by converting its bounding box sequence, $\{B_t = [x_{\min}, y_{\min}, x_{\max}, y_{\max}]\}_t$, into a structured string with explicit timestamps (e.g., \texttt{"t=1: [xmin, ymin, xmax, ymax]'; ..."}), making the raw kinematic information directly machine-readable.

\begin{figure}[h]
    \centering
    \includegraphics[width=0.8\columnwidth]{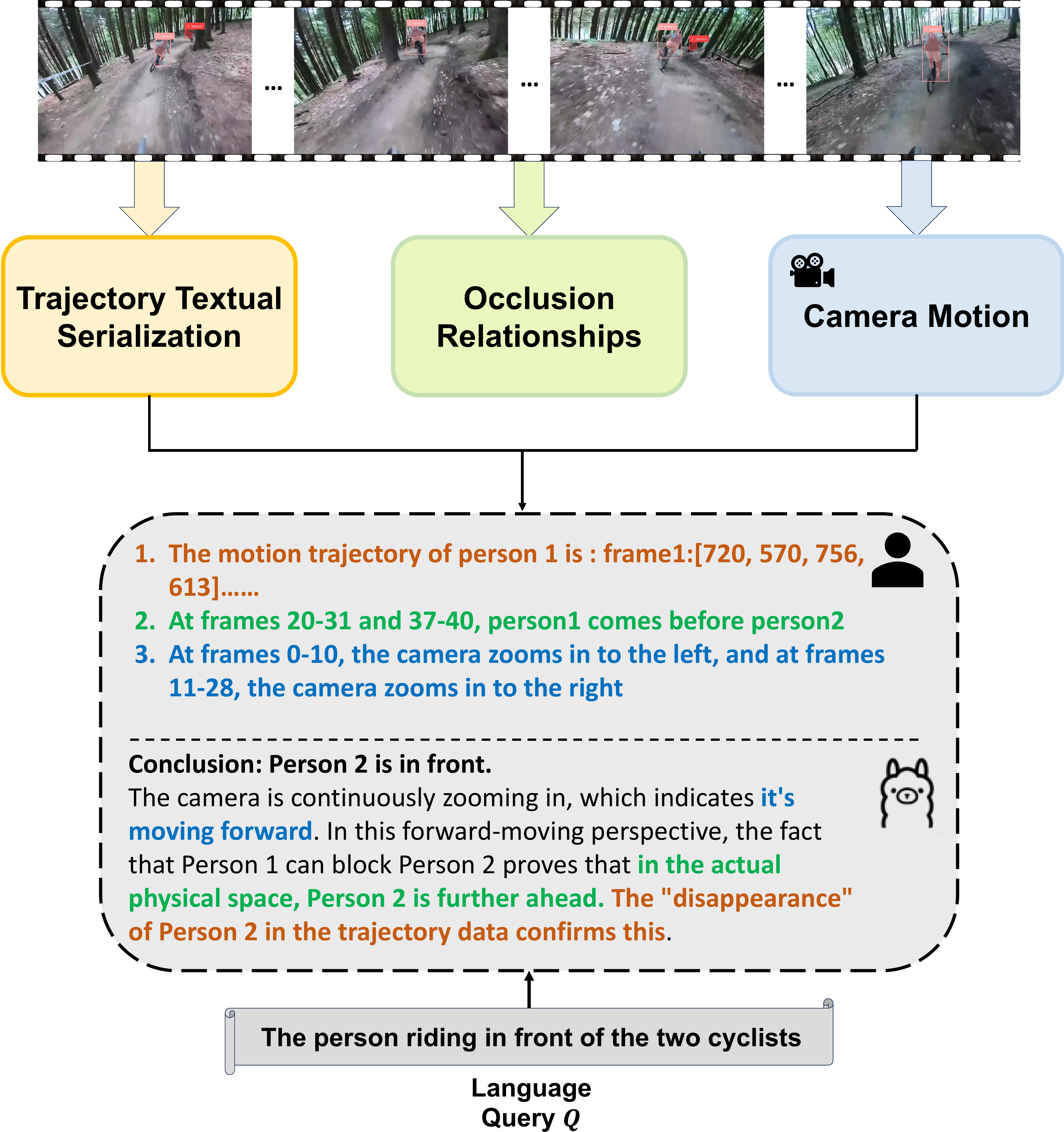}  
    \caption{Illustration of our coarse-grained motion reasoning. Given the query ``The person riding in front of the two cyclists", LLM is prompted with sequential texts, inferred occlusion relationships, and camera motion analysis.}
    \label{fig:llm_reasoning}
\end{figure}

Beyond the trajectory data itself, we provide two crucial contextual priors to enable more sophisticated reasoning. The first prior involves modeling \textbf{Occlusion Relationships} to imbue the model with an understanding of spatial hierarchy. This is inferred using a dynamic depth priority, where in any overlapping image region, the object with a larger pixel cardinality is assigned higher priority\cite{zhu2017semantic}. The second is a model of the \textbf{Camera Motion}, designed to help the LLM disentangle object movement from the viewpoint's own movement. This is achieved by estimating an inter-frame affine transformation matrix ($A_t$) via a robust sparse optical flow algorithm\cite{lucas1981iterative}, which minimizes the photometric error between frames:
\begin{equation}
A_t = \underset{A}{\operatorname{argmin}} \sum_i \left( I_{t-1}(\mathbf{x}_i) - I_t(A \cdot \mathbf{x}i) \right)^2,
\end{equation}
where $x_i$ represents a set of sparse feature points from frame $I_{t-1}$. Both priors are injected to the LLM as part of its prompt. With the availability of a complete prompt,  Llama 3 model is tasked to act as a zero-shot spatio-temporal reasoner. It synthesizes the three distinct streams of information, including the serialized trajectories, the camera motion model, and the occlusion data, to infer the intrinsic movement of each object. It then evaluates which candidate's behavior is most consistent with the user's motion query $Q_m$ to produce the filtered subset of trajectories $C'$. Figure \ref{fig:llm_reasoning} provides a concrete example of this complex reasoning process.

\subsubsection{Fine-grained Pose Verification}
This module is conditionally activated as the final arbiter when motion reasoning results in ambiguity (i.e., the number of candidates $|C'|$ is still greater than the expected target cardinality $K$) and a posture query $Q_p$ is available. Its purpose is to resolve such cases through fine-grained visual-semantic alignment. The process begins by selecting a small set of discriminative keyframes ($k=3$ in our experiments) where the candidates exhibit maximum visual separation, identified by finding the minimum Intersection over Union (IoU) between their bounding boxes. On these keyframes, we leverage CLIP to perform the matching. For each remaining candidate $c_i \in C'$, we crop its image region, pass it through CLIP's visual encoder $E_V$, and average the resulting $k$ feature vectors into a single aggregated visual embedding, $\bar{v}_i$. This embedding is then compared against the text embedding of the posture query, $v_p$, using cosine similarity to identify the definitive target, $c^*$, with the highest score:
\begin{equation}
    c^* = \operatorname*{argmax}_{c_i \in C'} \left( \frac{\bar{v}_i \cdot v_p}{\left\Vert \bar{v}_i \right\Vert \cdot \left\Vert v_p \right\Vert} \right).
\end{equation}
This fine-grained verification, focused purely on visual attributes, serves as the decisive step to pinpoint the correct target when motion cues are insufficient.

\begin{table*}[hbt]
\centering
\begin{tabular}{l | S | c c c | c c c | c c c} 
\hline
\multirow{2}{*}{Methods} & \multirow{2}{*}{Reference} & \multicolumn{3}{c|}{Ref-YouTube-VOS} & \multicolumn{3}{c|}{Ref-Davis17} & \multicolumn{3}{c}{MeviS} \\
& & $\mathcal{J}$ & $\mathcal{F}$ & $\mathcal{J}\&\mathcal{F}$ & $\mathcal{J}$ & $\mathcal{F}$ & $\mathcal{J}\&\mathcal{F}$ & $\mathcal{J}$ & $\mathcal{F}$ & $\mathcal{J}\&\mathcal{F}$ \\
\hline
ReferFormer~\citeyear{wu2022language} & [CVPR'22] & 61.3 & 64.6 & 62.9 & 58.1 & 64.1 & 61.1& 29.8 & 32.2 & 31.0 \\
SgMg~\citeyear{miao2023spectrum} & [ICCV'23] & 63.9 & 67.4 & 65.7 & 60.6 & 66.0 & 63.3 & - & - & - \\
LMPM~\citeyear{ding2023mevis} & [ICCV'23] & - & - & - & - & - & - & 34.2 & 40.2 & 37.2 \\
DsHmp~\citeyear{he2024decoupling} & [CVPR'24] & 65.0 & 69.1 & 67.1 & 61.7 & 68.1 & 64.9 & 43.0 & 49.8 & 46.4 \\
SSA~\citeyear{pan2025semantic} & [CVPR'25] & 62.2 & 66.4 & 64.3 & 64.0 & 70.7 & 67.3 & 44.3 & 53.4 & 48.9 \\
\hline
\multicolumn{11}{c}{\textit{With LLM || VLLM}} \\
\hline
Trackgpt-13B~\citeyear{zhu2023tracking} & [arXiv'23] & 58.1 & 60.8 & 59.5 & 62.7 & 70.4 & 66.5 & - & - & - \\
Vidoe-Lisa~\citeyear{bai2024one} & [NIPS'24] & 61.7 & 65.7 & 63.7 & 68.8 & 72.7 & 71.5 & 41.3 & 47.6 & 44.4 \\
VISA~\citeyear{yan2024visa} & [ECCV'24] & 61.4 & 64.7 & 63.0 & 67.0 & 73.8 & 70.4 & 41.8 & 47.1 & 44.5 \\
ViLLA~\citeyear{zheng2024villa} & [ICCV'25] & 64.6 & 70.4 & 67.5 & 70.6 & 78.0 & 74.3 & 46.5 & 52.3 & 49.4 \\
AL-Ref-SAM2*~\citeyear{huang2025unleashing} & [AAAI'25] & 65.9 & 69.9 & 67.9 & 70.4 & 78.0 & 74.2 & 39.5 & 46.2 & 42.8 \\
Glus~\citeyear{lin2025glus} & [arXiv'25] & 66.6 & 68.3 & 65.0 & - & - & - & 47.5 & 53.2 & 50.3 \\
VRS-HQ-13B~\citeyear{gong2025devil} & [CVPR'25] & 69.0 & 73.1 & 71.0 & 71.0 & 77.9 & 74.4 & 48.0 & 53.7 & 50.9 \\
\textbf{PARSE-VOS*(Ours)} & - & \textbf{70.0} & \textbf{74.2} & \textbf{72.1} & \textbf{72.4} & \textbf{78.6} & \textbf{75.5} & \textbf{49.2} & \textbf{55.6} & \textbf{52.4} \\
\hline
\end{tabular}
\caption{ Comparison with state-of-the-art methods on the validation sets of Ref-YouTube-VOS, Ref-DAVIS17 and MeViS
datasets. * Indicates that this is a training-free method.}
\label{tab:final_centered_comparison}
\end{table*}

\section{Experiments}
\subsection{Datasets and Evaluation Metrics}
\subsubsection{Dataset}
To evaluate the effectiveness and generalization capabilities of our proposed PARSE-VOS framework, we conducted experiments on three mainstream Referring Video Object Segmentation (RVOS) benchmarks, each with distinct characteristics. Our core evaluation is centered on the highly challenging \textbf{MeViS} dataset\cite{ding2023mevis}, which comprises 2,006 videos. Its scenarios feature multiple similar objects and complex motion patterns, specifically designed to test a model's spatio-temporal reasoning abilities. Additionally, we conducted tests on the largest-scale dataset, \textbf{Ref-YouTube-VOS}\cite{seo2020urvos},which contains 3,978 videos, to assess the model's performance in more general scenarios with relatively simple language descriptions. Finally, we utilized the \textbf{Ref-DAVIS17} dataset\cite{khoreva2018video}, which contains 90 videos, renowned for its high-quality pixel-level annotations, to precisely measure the model's segmentation accuracy when handling diverse objects.

\subsubsection{Evaluation Metrics}
We adopt the standard evaluation metrics\cite{khoreva2018video,seo2020urvos,ding2023mevis} in the field of Referring Video Object Segmentation (RVOS) to measure the performance of our model. These metrics include: \textbf{Region Similarity} ($\mathcal{J}$), which measures the Intersection over Union (IoU) between the predicted and ground-truth masks; \textbf{Contour Accuracy} ($\mathcal{F}$), which evaluates the boundary alignment; and their Average ($\mathcal{J}$ \& $\mathcal{F}$), which serves as the primary metric for overall performance comparison. All evaluations are conducted using the official evaluation code to ensure a fair comparison with previous works.

\subsection{Implementation Details}
All our experiments are conducted on a single NVIDIA RTX 4090 GPU. The language understanding and spatio-temporal reasoning tasks within our framework are handled by the Llama-3-8B-Instruct model \cite{meta2024llama3}. For spatio-temporal candidate grounding, we employ the lightweight GroundingDINO-T \cite{liu2023grounding} for object detection, with a box threshold of 0.3 and an IoU threshold of 0.4. We use SAM2-ViT-L \cite{ravi2024sam2segmentimages} for instance mask generation, with a keyframe sampling interval set to $\tau$=15. When fine-grained pose verification is necessary, the module selects k=3 of the most discriminative keyframes and utilizes the CLIP-ViT-L/14 \cite{clip} model to perform the final visual-semantic alignment to pinpoint the target.

\subsection{Comparison with State-of-the-Art Methods}
The proposed PARSE-VOS framework was compared against state-of-the-art methods on three widely used benchmarks, and the results are presented in Table 1.

MeViS is a highly challenging dataset characterized by its complex scenarios, which places high demands on a model's discriminative capabilities. As shown in Table \ref{tab:final_centered_comparison}, our method achieved a $\mathcal{J}\&\mathcal{F}$ score of 52.4\%. The result surpasses the leading non-LLM state-of-the-art method, SSA, by 3.5\% points and outperforms the leading LLM-based method, VRS-HQ-13B, by 1.5\% points. The performance on MeVis dataset validates the effectiveness of our proposed hierarchical reasoning framework in handling complex scenes with significant ambiguity.

On the largest-scale Ref-YouTube-VOS dataset, our model also demonstrated strong generalization capabilities. This dataset covering general scenarios with relatively simple language descriptions. PARSE-VOS achieved a $\mathcal{J}\&\mathcal{F}$ score of 72.1\%, ranking first among all compared methods.

To measure the model's segmentation accuracy, we conducted experiments on the Ref-DAVIS17 dataset, which is known for its high-quality pixel-level annotations. Our method achieved a $\mathcal{J}\&\mathcal{F}$ score of 75.5\% on this dataset, demonstrating its capability to generate high-precision masks. A noteworthy issue is that our framework achieved this outstanding performance while utilizing a relatively compact Llama-3-8B-Instruct model. As shown in Table \ref{tab:final_centered_comparison}, our method outperforms approaches that rely on larger-scale models, such as AL-Ref-SAM2\cite{huang2025unleashing} (gpt4), Trackgpt-13B\cite{zhu2023tracking} and VRS-HQ-13B\cite{gong2025devil}. This demonstrates the superiority of the proposed hierarchical coarse-to-fine reasoning architecture over the original large language models.

\begin{figure*}[t]
    \centering
    \includegraphics[width=1.0\textwidth]{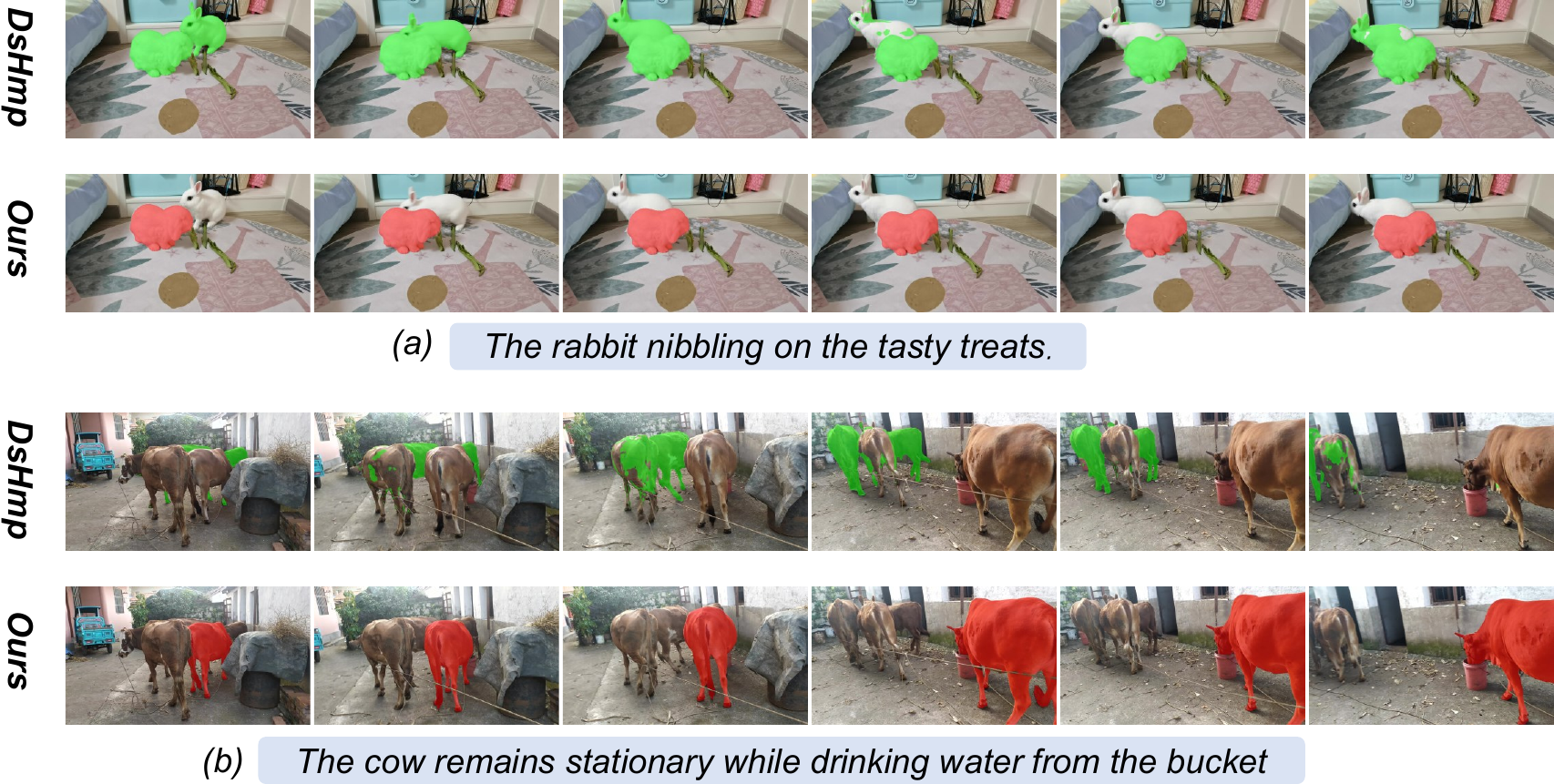}
    \caption{Qualitative results on the MeViS dataset, showing our framework's robustness and advanced contextual reasoning. (a) This demonstrates robustness to interference from similar objects. Unlike the baseline, our model precisely segments the target rabbit while ignoring the distractor. (b) This highlights advanced contextual reasoning. By analyzing camera motion, it uses motion matching and the contextual “bucket” to accurately identify the stationary cow from a complex query.}
    \label{fig:experiment_figure}
\end{figure*}

\subsection{Ablative Study}
\subsubsection{Analysis of the Hierarchical Reasoning Architecture}

To validate the effectiveness of our hierarchical reasoning architecture, we conduct several additive ablation experiments. Results are presented in Table \ref{tab:your_imitated_style_table}. The baseline model removes the entire hierarchical reasoning for target identification module but randomly selects $K$ targets from all candidate trajectories. This naive approach achieves a $\mathcal{J}\&\mathcal{F}$ score of only 33.7\%.
On this basis, individually introducing the Coarse-grained Motion Reasoning (CMR) or the Fine-grained Pose Verification (FPV) module substantially boosts the $\mathcal{J}\&\mathcal{F}$ score to 47.0\% (+13.3) and 39.4\% (+5.7), respectively. This demonstrates the effectiveness of each component, with the CMR module making the most significant contribution. Finally, when both modules are cascaded to form our full model, the performance reaches an optimal 52.4\%. This result, surpassing any single component, clearly validates the efficiency and complementarity of our ``coarse-to-fine" hierarchical design: CMR performs initial coarse-grained filtering, while FPV provides subsequent fine-grained judgment, leading to the best performance.

\begin{table}[t]
\centering
\begin{tabular}{l|ccc}
\hline
\multirow{2}{*}{Method} & \multicolumn{3}{c}{MeViS} \\
& $\mathcal{J}$ & $\mathcal{F}$ & $\mathcal{J}\&\mathcal{F}$ \\
\hline
baseline         & 32.1 & 35.3 & 33.7 \\
+CMR     & 45.8 & 48.2 & 47.0 \\
+FPV     & 38.4 & 40.3 & 39.4 \\
+CMR+FPV & 49.2 & 55.6 & 52.4 \\
\hline
\end{tabular}
\caption{Additive Ablation Study of the Hierarchical Reasoning Architecture on the MeViS dataset}
\label{tab:your_imitated_style_table}
\end{table}

\subsubsection{Analysis of Contextual Priors in Coarse-grained Motion Reasoning}

\begin{table}[t]
\centering
\begin{tabular}{l|ccc}
\hline
\multicolumn{1}{c}{\multirow{2}{*}{Method}} & \multicolumn{3}{c}{MeViS} \\
& $\mathcal{J}$ & $\mathcal{F}$ & $\mathcal{J}\&\mathcal{F}$ \\
\hline
Trajectory Only & 42.8 & 45.8 & 44.3 \\
\quad + CMM & 45.2 & 50.3 & 47.8 \\
\quad + OR & 44.9 & 48.5 & 46.7 \\
\hline
\textbf{Full Context} & \textbf{49.2} & \textbf{55.6} & \textbf{52.4} \\
\hline
\end{tabular}
\caption{Additive contribution from contextual priors (CMM \& OR) on the basis of a trajectory-only reasoning model.}
\label{tab:ablation_context_final}
\end{table}

To quantify the contribution of our contextual priors, we conducted an ablation study on the inputs to the coarse-grained motion reasoning module. As shown in Table \ref{tab:ablation_context_final}, we analyze the model's performance when we selectively evaluate the \textbf{Camera Motion Model (CMM)} and \textbf{Occlusion Relationships (OR)}. Our analysis starts with the Trajectory Only configuration, which refers to our full framework except the reasoning module is provided only with serialized trajectory data without any additional context. This baseline configuration achieves a $\mathcal{J}\&\mathcal{F}$ score of 44.3\%. Building on this, individually supplementing the reasoner with \textbf{CMM} or \textbf{OR} brings steady performance gains, reaching 47.8\% and 46.7\% respectively. Finally, when the full context is provided, that is, when the reasoner utilizes trajectory data, CMM and OR together, the performance is optimal at 52.4\%. This clearly demonstrates that both CMM and OR are crucial, complementary components that enable the LLM to perform sophisticated spatio-temporal reasoning beyond simple coordinate matching.

\subsubsection{Qualitative Results}

Qualitative comparisons on the challenging MeViS dataset are presented in Fig. \ref{fig:experiment_figure}. In scenario (a), DsHmp\cite{he2024decoupling} is clearly confused by a similar distractor rabbit, whereas our method segments the target precisely. In scenario (b), which requires more complex reasoning, our method excels by parsing and utilizing subtle motion cues (``remains stationary") and contextual information (``bucket") from the language query, and thus can accurately identify the target cow. These results demonstrate the robustness of our hierarchical reasoning framework in handling high-ambiguity scenarios.

\section{Conclusion}
We present PARSE-VOS, a training-free, hierarchical reasoning framework for Referring Video Object Segmentation (RVOS). Our method performs target identification through a two-stage, coarse-to-fine reasoning process, enriching large language model (LLM) decisions with contextual priors such as camera motion. PARSE-VOS achieves state-of-the-art results across three major RVOS benchmarks. Notably, our compact 8B model outperforms larger-scale counterparts, highlighting that effective reasoning architecture can be more impactful than simply scaling model size.

\bibliography{references}

\begin{thebibliography}{37}
\providecommand{\natexlab}[1]{#1}

\bibitem[{Bai et~al.(2024)Bai, He, Mei, Wang, Gao, Chen, Liu, Zhang, and Shou}]{bai2024one}
Bai, Z.; He, T.; Mei, H.; Wang, P.; Gao, Z.; Chen, J.; Liu, L.; Zhang, Z.; and Shou, M.~Z. 2024.
\newblock One Token to Seg Them All: Language Instructed Reasoning Segmentation in Videos.
\newblock In \emph{Advances in Neural Information Processing Systems}.

\bibitem[{Bar-Tal et~al.(2022)Bar-Tal, Ofri-Amar, Fridman, Kasten, and Dekel}]{Bar-Tal2022Text2live}
Bar-Tal, O.; Ofri-Amar, D.; Fridman, R.; Kasten, Y.; and Dekel, T. 2022.
\newblock Text2live: Text-driven layered image and video editing.
\newblock In \emph{European Conference on Computer Vision(ECCV)}, volume 13675 of \emph{Lecture Notes in Computer Science}, 707--723. Springer Nature Switzerland.

\bibitem[{Botach, Zheltonozhskii, and Baskin(2022)}]{botach2022end}
Botach, A.; Zheltonozhskii, E.; and Baskin, C. 2022.
\newblock End-to-end referring video object segmentation with multimodal transformers.
\newblock In \emph{Proceedings of the IEEE/CVF Conference on Computer Vision and Pattern Recognition (CVPR)}, 19777--19787.

\bibitem[{Ding et~al.(2023)Ding, Liu, He, Jiang, and Loy}]{ding2023mevis}
Ding, H.; Liu, C.; He, S.; Jiang, X.; and Loy, C.~C. 2023.
\newblock {MeViS}: A Large-scale Benchmark for Video Segmentation with Motion Expressions.
\newblock In \emph{Proceedings of the IEEE/CVF International Conference on Computer Vision}, 10156--10166.

\bibitem[{Gao et~al.(2025{\natexlab{a}})Gao, Wu, Zhang, Tian, Zhou, and Tu}]{Gao2025Automated}
Gao, X.; Wu, K.; Zhang, H.; Tian, K.; Zhou, Y.; and Tu, Z. 2025{\natexlab{a}}.
\newblock Automated Vehicles Should be Connected with Natural Language.
\newblock \emph{arXiv preprint arXiv:2507.01059}.

\bibitem[{Gao et~al.(2025{\natexlab{b}})Gao, Wu, Wang, Liu, Zhou, and Tu}]{Gao2025LangCoop}
Gao, X.; Wu, Y.; Wang, R.; Liu, C.; Zhou, Y.; and Tu, Z. 2025{\natexlab{b}}.
\newblock LangCoop: Collaborative Driving with Language.
\newblock \emph{arXiv preprint arXiv:2504.13406}.

\bibitem[{Gavrilyuk et~al.(2017)Gavrilyuk, Ghodrati, Li, and Snoek}]{Gavrilyuk2017Actor}
Gavrilyuk, K.; Ghodrati, A.; Li, Z.; and Snoek, C. G.~M. 2017.
\newblock Actor and Action Video Segmentation from a Sentence.
\newblock In \emph{Proceedings of the IEEE/CVF Conference on Computer Vision and Pattern Recognition (CVPR)}, 5958--5966.

\bibitem[{Gong et~al.(2024)Gong, Lin, Zhang, Lu, Han, Liu, Guo, Lin, and Wan}]{guiji2}
Gong, L.; Lin, Y.; Zhang, X.; Lu, Y.; Han, X.; Liu, Y.; Guo, S.; Lin, Y.; and Wan, H. 2024.
\newblock Mobility-{LLM}: Learning Visiting Intentions and Travel Preferences from Human Mobility Data with Large Language Models.
\newblock In \emph{Thirty-eighth Conference on Neural Information Processing Systems}.

\bibitem[{Gong et~al.(2025)Gong, Zhuge, Zhang, Yang, Zhang, and Lu}]{gong2025devil}
Gong, S.; Zhuge, Y.; Zhang, L.; Yang, Z.; Zhang, P.; and Lu, H. 2025.
\newblock The Devil is in Temporal Token: High Quality Video Reasoning Segmentation.
\newblock In \emph{Proceedings of the IEEE/CVF Conference on Computer Vision and Pattern Recognition (CVPR)}.

\bibitem[{He and Ding(2024)}]{he2024decoupling}
He, S.; and Ding, H. 2024.
\newblock Decoupling static and hierarchical motion perception for referring video segmentation.
\newblock In \emph{CVPR}.

\bibitem[{Huang et~al.(2025)Huang, Ling, Li, Hui, Tang, Wei, Han, and Liu}]{huang2025unleashing}
Huang, S.; Ling, R.; Li, H.; Hui, T.; Tang, Z.; Wei, X.; Han, J.; and Liu, S. 2025.
\newblock Unleashing the Temporal-Spatial Reasoning Capacity of GPT for Training-Free Audio and Language Referenced Video Object Segmentation.
\newblock In \emph{Proceedings of the AAAI Conference on Artificial Intelligence}.

\bibitem[{Khoreva, Rohrbach, and Schiele(2018)}]{khoreva2018video}
Khoreva, A.; Rohrbach, A.; and Schiele, B. 2018.
\newblock Video Object Segmentation with Language Referring Expressions.
\newblock In \emph{Proceedings of the Asian Conference on Computer Vision (ACCV)}.

\bibitem[{Kim et~al.(2025)Kim, Jin, Choi, Lim, Kim, and Yoon}]{Kim2025SOLA}
Kim, S.; Jin, W.; Choi, H.; Lim, S.; Kim, S.; and Yoon, H. 2025.
\newblock Referring Video Object Segmentation via Language-aligned Track Selection.
\newblock \emph{arXiv preprint arXiv:2412.01136}.

\bibitem[{Li et~al.(2023)Li, Wang, Xu, Li, Raj, and Lu}]{Li2023Robust}
Li, X.; Wang, J.; Xu, X.; Li, X.; Raj, B.; and Lu, Y. 2023.
\newblock Robust Referring Video Object Segmentation with Cyclic Structural Consensus.
\newblock In \emph{Proceedings of the IEEE/CVF International Conference on Computer Vision (ICCV)}, 22236--22245.

\bibitem[{Liang et~al.(2025)Liang, Lin, Tan, Zhang, Zheng, and Hu}]{liang2025referdino}
Liang, T.; Lin, K.-Y.; Tan, C.; Zhang, J.; Zheng, W.-S.; and Hu, J.-F. 2025.
\newblock ReferDINO: Referring Video Object Segmentation with Visual Grounding Foundations.
\newblock In \emph{Proceedings of the IEEE/CVF International Conference on Computer Vision}.

\bibitem[{Lin et~al.(2025)Lin, Wang, Yu, and Pang}]{lin2025glus}
Lin, L.; Wang, Y.-X.; Yu, X.; and Pang, Z. 2025.
\newblock {GLUS: Global-Local Reasoning Unified into A Single Large Language Model for Video Segmentation}.
\newblock \emph{arXiv preprint arXiv:2504.07962}.

\bibitem[{Liu, Wu, and Xie(2024)}]{Liu-PRAI-2024}
Liu, D.; Wu, L.~Y.; and Xie, X. 2024.
\newblock Blended Latent Diffusion under Attention Control for Real-World Video Editing.
\newblock In \emph{International Conference on Pattern Recognition and Artificial Intelligence}, --.

\bibitem[{Liu et~al.(2023)Liu, Zeng, Ren, Li, Zhang, Yang, Li, Yang, Su, Zhu et~al.}]{liu2023grounding}
Liu, S.; Zeng, Z.; Ren, T.; Li, F.; Zhang, H.; Yang, J.; Li, C.; Yang, J.; Su, H.; Zhu, J.; et~al. 2023.
\newblock Grounding dino: Marrying dino with grounded pre-training for open-set object detection.
\newblock \emph{arXiv preprint arXiv:2303.05499}.

\bibitem[{Lucas and Kanade(1981)}]{lucas1981iterative}
Lucas, B.~D.; and Kanade, T. 1981.
\newblock {An Iterative Image Registration Technique with an Application to Stereo Vision}.
\newblock In \emph{Proceedings of the 7th International Joint Conference on Artificial Intelligence (IJCAI '81)}, 674--679.

\bibitem[{{Meta AI}(2024)}]{meta2024llama3}
{Meta AI}. 2024.
\newblock {The Llama 3 Herd of Models}.
\newblock arXiv:2404.11225.

\bibitem[{Miao et~al.(2023)Miao, Bennamoun, Gao, and Mian}]{miao2023spectrum}
Miao, B.; Bennamoun, M.; Gao, Y.; and Mian, A. 2023.
\newblock Spectrum-guided multi-granularity referring video object segmentation.
\newblock In \emph{ICCV}.

\bibitem[{Pan et~al.(2025)Pan, Fang, Li, Xu, Li, Benini, and Lu}]{pan2025semantic}
Pan, F.; Fang, H.; Li, F.; Xu, Y.; Li, Y.; Benini, L.; and Lu, X. 2025.
\newblock Semantic and Sequential Alignment for Referring Video Object Segmentation.
\newblock In \emph{Proceedings of the IEEE/CVF Conference on Computer Vision and Pattern Recognition (CVPR)}.

\bibitem[{Qi et~al.(2020)Qi, Wu, Anderson, Wang, Wang, Shen, and van~den Hengel}]{Qi2020Reverie}
Qi, Y.; Wu, Q.; Anderson, P.; Wang, X.; Wang, W.~Y.; Shen, C.; and van~den Hengel, A. 2020.
\newblock {REVERIE}: Remote Embodied Visual Referring Expression in Real Indoor Environments.
\newblock In \emph{Proceedings of the {IEEE/CVF} Conference on Computer Vision and Pattern Recognition (CVPR)}, 9985--9994.

\bibitem[{Qin et~al.(2023)Qin, Han, Wang, Nie, Yin, and Xiankai}]{Qin2023Unified}
Qin, Z.; Han, C.; Wang, Q.; Nie, X.; Yin, Y.; and Xiankai, L. 2023.
\newblock Unified {3D} Segmenter as Prototypical Classifiers.
\newblock In \emph{Advances in Neural Information Processing Systems (NeurIPS)}, volume~36, 5315--5328.

\bibitem[{Radford et~al.(2021)Radford, Kim, Hallacy, Ramesh, Goh, Agarwal, Sastry, Askell, Mishkin, Clark, Krueger, and Sutskever}]{clip}
Radford, A.; Kim, J.~W.; Hallacy, C.; Ramesh, A.; Goh, G.; Agarwal, S.; Sastry, G.; Askell, A.; Mishkin, P.; Clark, J.; Krueger, G.; and Sutskever, I. 2021.
\newblock {Learning Transferable Visual Models From Natural Language Supervision}.
\newblock In \emph{Proceedings of the 38th International Conference on Machine Learning (ICML)}, volume 139 of \emph{Proceedings of Machine Learning Research}, 8748--8763. PMLR.

\bibitem[{Ravi et~al.(2024)Ravi, Gabeur, Hu, Hu, Ryali, Ma, Khedr, Rädle, Rolland, Gustafson, Mintun, Pan, Alwala, Carion, Wu, Girshick, Dollár, and Feichtenhofer}]{ravi2024sam2segmentimages}
Ravi, N.; Gabeur, V.; Hu, Y.-T.; Hu, R.; Ryali, C.; Ma, T.; Khedr, H.; Rädle, R.; Rolland, C.; Gustafson, L.; Mintun, E.; Pan, J.; Alwala, K.~V.; Carion, N.; Wu, C.-Y.; Girshick, R.; Dollár, P.; and Feichtenhofer, C. 2024.
\newblock SAM 2: Segment Anything in Images and Videos.
\newblock arXiv:2408.00714.

\bibitem[{Seo, Lee, and Han(2020)}]{seo2020urvos}
Seo, S.; Lee, J.-Y.; and Han, B. 2020.
\newblock {URVOS}: Unified Referring Video Object Segmentation Network with a Large-Scale Benchmark.
\newblock In \emph{Proceedings of the European Conference on Computer Vision (ECCV)}, 218--234.

\bibitem[{Vaswani et~al.(2017)Vaswani, Shazeer, Parmar, Uszkoreit, Jones, Gomez, Kaiser, and Polosukhin}]{transfomer}
Vaswani, A.; Shazeer, N.; Parmar, N.; Uszkoreit, J.; Jones, L.; Gomez, A.~N.; Kaiser, {\L}.; and Polosukhin, I. 2017.
\newblock Attention is all you need.
\newblock In \emph{Advances in neural information processing systems}, 5998--6008.

\bibitem[{Wu et~al.(2022)Wu, Jiang, Sun, Yuan, and Luo}]{wu2022language}
Wu, J.; Jiang, Y.; Sun, P.; Yuan, Z.; and Luo, P. 2022.
\newblock Language as queries for referring video object segmentation.
\newblock In \emph{CVPR}.

\bibitem[{Yan et~al.(2024)Yan, Wang, Yan, Jiang, Hu, Kang, Xie, and Gavves}]{yan2024visa}
Yan, C.; Wang, H.; Yan, S.; Jiang, X.; Hu, Y.; Kang, G.; Xie, W.; and Gavves, E. 2024.
\newblock VISA: Reasoning Video Object Segmentation via Large Language Models.
\newblock \emph{arXiv preprint arXiv:2407.11325}.

\bibitem[{Yang et~al.(2025)Yang, Guo, Lin, Dong, Huang, Wu, Zuo, Peng, Zhong, Wang, Guo, Jia, Yan, and Lin}]{guiji3}
Yang, K.; Guo, Z.; Lin, G.; Dong, H.; Huang, Z.; Wu, Y.; Zuo, D.; Peng, J.; Zhong, Z.; Wang, X.; Guo, Q.; Jia, X.; Yan, J.; and Lin, D. 2025.
\newblock {TRAJECTORY-LLM}: A Language-Based Data Generator for Trajectory Prediction in Autonomous Driving.
\newblock In \emph{International Conference on Learning Representations}.

\bibitem[{Ying et~al.(2023)Ying, Zhong, Mao, Wu, Chen, Liu, Fan, Zhuge, and Shen}]{CTVIS-2023}
Ying, K.; Zhong, Q.; Mao, W.; Wu, L.~Y.; Chen, H.; Liu, Y.; Fan, C.; Zhuge, Y.; and Shen, C. 2023.
\newblock CTVIS: Consistent Training for Online Video Instance Segmentation.
\newblock In \emph{IEEE International Conference on Computer Vision (ICCV)}, --.

\bibitem[{Zhang et~al.(2023)Zhang, Amiri, Liu, Züfle, and Zhao}]{guiji1}
Zhang, Z.; Amiri, H.; Liu, Z.; Züfle, A.; and Zhao, L. 2023.
\newblock Large Language Models for Spatial Trajectory Patterns Mining.
\newblock arXiv:2310.04942.

\bibitem[{Zheng et~al.(2024)Zheng, Qi, Chen, Wang, Wang, Qiao, and Zhao}]{zheng2024villa}
Zheng, R.; Qi, L.; Chen, X.; Wang, Y.; Wang, K.; Qiao, Y.; and Zhao, H. 2024.
\newblock ViLLa: Video Reasoning Segmentation with Large Language Model.
\newblock \emph{arXiv preprint arXiv:2407.14500}.

\bibitem[{Zhong et~al.(2025)Zhong, Jiang, Wang, Ding, Wu, and Huang}]{ZhongICME-2025}
Zhong, Q.; Jiang, P.; Wang, W.; Ding, G.; Wu, L.~Y.; and Huang, K. 2025.
\newblock A Temporal Modeling Framework for Video Pre-Training on Video Instance Segmentation.
\newblock In \emph{International Conference on Multimedia Expo (ICME)}, --.

\bibitem[{Zhu et~al.(2023)Zhu, Cheng, He, Li, Luo, Lu, Geng, and Xie}]{zhu2023tracking}
Zhu, J.; Cheng, Z.-Q.; He, J.-Y.; Li, C.; Luo, B.; Lu, H.; Geng, Y.; and Xie, X. 2023.
\newblock Tracking with human-intent reasoning.
\newblock arXiv:2312.17448.

\bibitem[{Zhu et~al.(2017)Zhu, Tian, Li, and Metaxas}]{zhu2017semantic}
Zhu, Y.; Tian, Y.; Li, G.-T.; and Metaxas, D.~N. 2017.
\newblock Semantic Amodal Instance Segmentation.
\newblock In \emph{Proceedings of the IEEE International Conference on Computer Vision (ICCV)}, 2128--2136.

\end{thebibliography}
\newcommand{\isChecklistMainFile}{} 
\makeatletter
\@ifundefined{isChecklistMainFile}{
  \newif\ifreproStandalone
  \reproStandalonetrue
}{
  \newif\ifreproStandalone
  \reproStandalonefalse
}
\makeatother

\ifreproStandalone
\documentclass[letterpaper]{article}
\usepackage[submission]{aaai2026}
\setlength{\pdfpagewidth}{8.5in}
\setlength{\pdfpageheight}{11in}
\usepackage{times}
\usepackage{helvet}
\usepackage{courier}
\usepackage{xcolor}
\frenchspacing

\begin{document}
\fi
\setlength{\leftmargini}{20pt}
\makeatletter\def\@listi{\leftmargin\leftmargini \topsep .5em \parsep .5em \itemsep .5em}
\def\@listii{\leftmargin\leftmarginii \labelwidth\leftmarginii \advance\labelwidth-\labelsep \topsep .4em \parsep .4em \itemsep .4em}
\def\@listiii{\leftmargin\leftmarginiii \labelwidth\leftmarginiii \advance\labelwidth-\labelsep \topsep .4em \parsep .4em \itemsep .4em}\makeatother

\setcounter{secnumdepth}{0}
\renewcommand\thesubsection{\arabic{subsection}}
\renewcommand\labelenumi{\thesubsection.\arabic{enumi}}

\newcounter{checksubsection}
\newcounter{checkitem}[checksubsection]

\newcommand{\checksubsection}[1]{%
  \refstepcounter{checksubsection}%
  \paragraph{\arabic{checksubsection}. #1}%
  \setcounter{checkitem}{0}%
}

\newcommand{\checkitem}{%
  \refstepcounter{checkitem}%
  \item[\arabic{checksubsection}.\arabic{checkitem}.]%
}
\newcommand{\question}[2]{\normalcolor\checkitem #1 #2 \color{blue}}
\newcommand{\ifyespoints}[1]{\makebox[0pt][l]{\hspace{-15pt}\normalcolor #1}}

\section*{Reproducibility Checklist}

\vspace{1em}


\checksubsection{General Paper Structure}
\begin{itemize}

\question{Includes a conceptual outline and/or pseudocode description of AI methods introduced}{(yes/partial/no/NA)}
yes

\question{Clearly delineates statements that are opinions, hypothesis, and speculation from objective facts and results}{(yes/no)}
yes

\question{Provides well-marked pedagogical references for less-familiar readers to gain background necessary to replicate the paper}{(yes/no)}
yes

\end{itemize}
\checksubsection{Theoretical Contributions}
\begin{itemize}

\question{Does this paper make theoretical contributions?}{(yes/no)}
no

	\ifyespoints{\vspace{1.2em}If yes, please address the following points:}
        \begin{itemize}
	
	\question{All assumptions and restrictions are stated clearly and formally}{(yes/partial/no)}
	NA

	\question{All novel claims are stated formally (e.g., in theorem statements)}{(yes/partial/no)}
	NA

	\question{Proofs of all novel claims are included}{(yes/partial/no)}
	NA

	\question{Proof sketches or intuitions are given for complex and/or novel results}{(yes/partial/no)}
	NA

	\question{Appropriate citations to theoretical tools used are given}{(yes/partial/no)}
	NA

	\question{All theoretical claims are demonstrated empirically to hold}{(yes/partial/no/NA)}
	NA

	\question{All experimental code used to eliminate or disprove claims is included}{(yes/no/NA)}
	NA
	
	\end{itemize}
\end{itemize}

\checksubsection{Dataset Usage}
\begin{itemize}

\question{Does this paper rely on one or more datasets?}{(yes/no)}
yes

\ifyespoints{If yes, please address the following points:}
\begin{itemize}

	\question{A motivation is given for why the experiments are conducted on the selected datasets}{(yes/partial/no/NA)}
	yes

	\question{All novel datasets introduced in this paper are included in a data appendix}{(yes/partial/no/NA)}
	NA

	\question{All novel datasets introduced in this paper will be made publicly available upon publication of the paper with a license that allows free usage for research purposes}{(yes/partial/no/NA)}
	NA

	\question{All datasets drawn from the existing literature (potentially including authors' own previously published work) are accompanied by appropriate citations}{(yes/no/NA)}
	yes

	\question{All datasets drawn from the existing literature (potentially including authors' own previously published work) are publicly available}{(yes/partial/no/NA)}
	yes

	\question{All datasets that are not publicly available are described in detail, with explanation why publicly available alternatives are not scientifically satisficing}{(yes/partial/no/NA)}
	NA

\end{itemize}
\end{itemize}

\checksubsection{Computational Experiments}
\begin{itemize}

\question{Does this paper include computational experiments?}{(yes/no)}
yes

\ifyespoints{If yes, please address the following points:}
\begin{itemize}

	\question{This paper states the number and range of values tried per (hyper-) parameter during development of the paper, along with the criterion used for selecting the final parameter setting}{(yes/partial/no/NA)}
	yes

	\question{Any code required for pre-processing data is included in the appendix}{(yes/partial/no)}
	yes

	\question{All source code required for conducting and analyzing the experiments is included in a code appendix}{(yes/partial/no)}
	yes

	\question{All source code required for conducting and analyzing the experiments will be made publicly available upon publication of the paper with a license that allows free usage for research purposes}{(yes/partial/no)}
	yes
        
	\question{All source code implementing new methods have comments detailing the implementation, with references to the paper where each step comes from}{(yes/partial/no)}
	yes

	\question{If an algorithm depends on randomness, then the method used for setting seeds is described in a way sufficient to allow replication of results}{(yes/partial/no/NA)}
	yes

	\question{This paper specifies the computing infrastructure used for running experiments (hardware and software), including GPU/CPU models; amount of memory; operating system; names and versions of relevant software libraries and frameworks}{(yes/partial/no)}
	yes

	\question{This paper formally describes evaluation metrics used and explains the motivation for choosing these metrics}{(yes/partial/no)}
	yes

	\question{This paper states the number of algorithm runs used to compute each reported result}{(yes/no)}
	yes

	\question{Analysis of experiments goes beyond single-dimensional summaries of performance (e.g., average; median) to include measures of variation, confidence, or other distributional information}{(yes/no)}
	no

	\question{The significance of any improvement or decrease in performance is judged using appropriate statistical tests (e.g., Wilcoxon signed-rank)}{(yes/partial/no)}
	no

	\question{This paper lists all final (hyper-)parameters used for each model/algorithm in the paper’s experiments}{(yes/partial/no/NA)}
	yes

\end{itemize}
\end{itemize}
\ifreproStandalone
\end{document}
\fi
\clearpage
\section{Appendix}
\subsection{A. LLM Prompts and Hyperparameter Settings}
\begin{figure*}[t]
    \centering
    \includegraphics[width=0.78\textwidth]{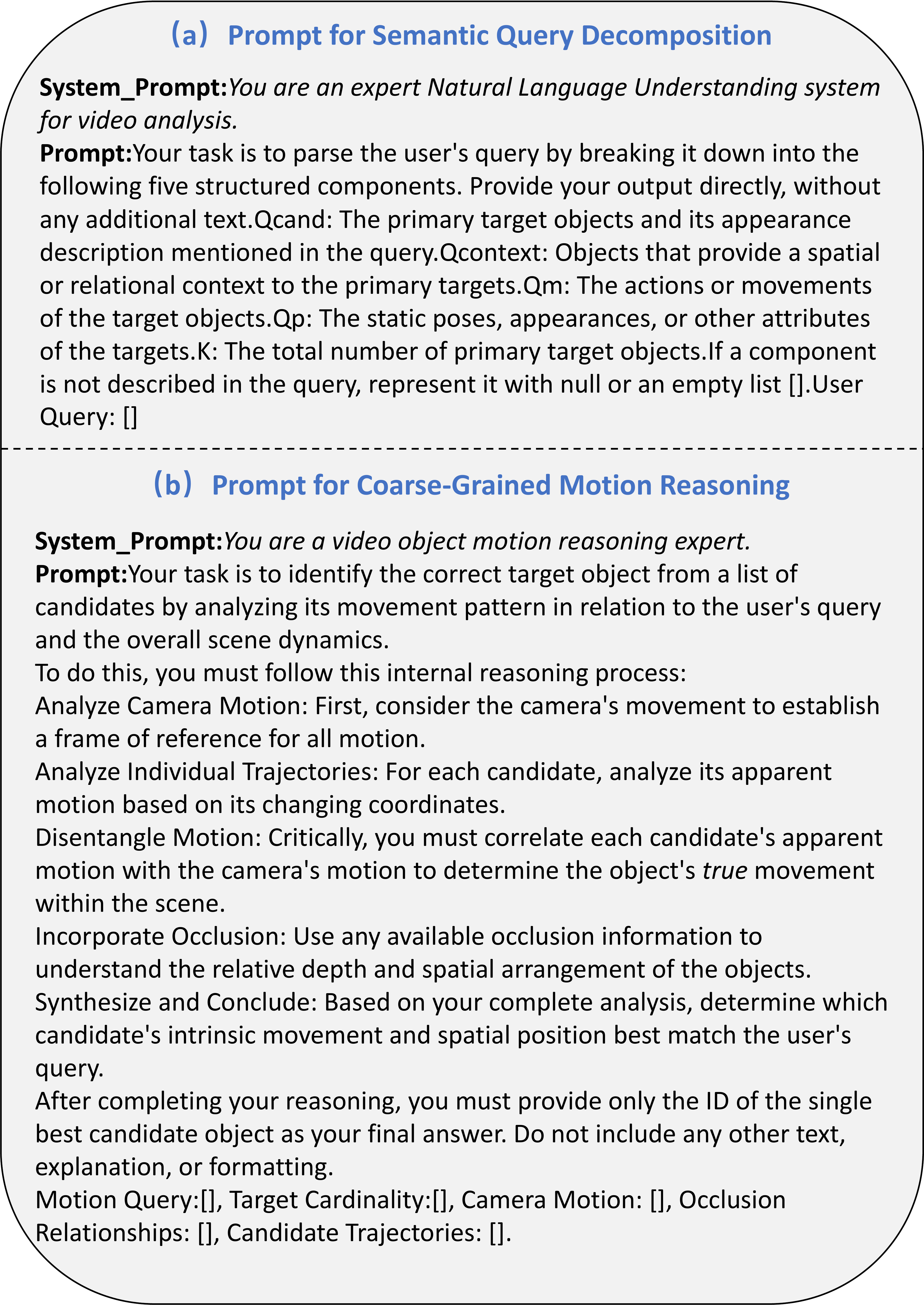}
    \caption{LLM Prompts for Query Decomposition and Motion Reasoning. The figure presents the specific prompts used in our framework. (a) Prompt for Semantic Query Decomposition: A prompt designed to transform a natural language query into a structured, five-part command. (b) Prompt for Coarse-Grained Motion Reasoning: A prompt that guides an LLM to identify a target object by following a strict five-step reasoning process. It synthesizes trajectory, camera motion, and occlusion data to output a single, definitive target ID.}
    \label{fig:prompt_figure}
\end{figure*}

This section provides the complete prompts and hyperparameter configurations for the Large Language Model (LLM) modules within our PARSE-VOS framework, ensuring the full reproducibility of our results. All experiments were conducted using the Llama-3-8B-Instruct model\cite{meta2024llama3}.

For all generative tasks performed by the LLM, we employed a consistent set of decoding parameters to ensure stable and high-quality outputs. The hyperparameter settings are as follows:
\begin{itemize}
    \item temperature = 0.7: This value was chosen to maintain a balance between deterministic accuracy for the parsing task and sufficient creativity for the complex reasoning task.
    
    \item top\_p = 0.95: We use nucleus sampling to ensure that the model's output remains coherent and high-quality by sampling from the most probable tokens that constitute a cumulative probability of 95\%.
    
    \item top\_k = 0: This setting disables top-k sampling, allowing nucleus sampling (top\_p) to be the primary method for controlling token selection.
\end{itemize}

Fig. \ref{fig:prompt_figure} illustrates the exact prompts provided to the LLM. These prompts were carefully engineered to guide the model's behavior for the two key tasks in our pipeline: (a) parsing the natural language query into structured commands , and (b) performing coarse-grained motion reasoning by synthesizing trajectory data, camera motion, and occlusion information.

\subsection{B. Parameter Analysis}
We conduct detailed ablation studies on key hyperparameters to validate the settings used in our framework. These analyses demonstrate the rationale behind our choices by exploring the trade-off between performance and efficiency.

\paragraph{Keyframe Sampling Interval ($\tau$).}
The parameter $\tau$ determines the sampling interval for keyframes where instance segmentation is performed. A careful balance is required. As shown in Table \ref{tab:pai_figre}, setting $\tau$ too low (e.g., 5 or 10) not only increases computational cost but can also introduce tracking errors, especially for slow-moving or crowded objects where minimal changes between frames might confuse the association module. Conversely, setting $\tau$ too high (e.g., 20 or 30) can cause the model to completely miss objects or significant events that occur between the sparse keyframes. The results clearly indicate that the model achieves its peak performance with a $\mathcal{J}\&\mathcal{F}$ score of 52.4 when $\tau$=15. Thus, we adopt $\tau$=15 as the default setting in our experiments.

\paragraph{Verification Window Size.}
For our predictive association criterion, the size of the Verification Window determines the number of frames to propagate forward to robustly match instance trajectories. As shown in Table \ref{tab:k_figre}, when the window size is 1, our predictive criterion degenerates into a less robust, single-frame similarity check only based on IoU, failing to leverage trajectory information. Increasing the window size to 3 yields a significant performance boost (+5.7 $\mathcal{J}\&\mathcal{F}$). While a larger window of 15 frames provides a marginal further improvement, it incurs a substantial computational cost from the increased mask propagation operations. Therefore, to optimize the trade-off between association accuracy and computational efficiency, we select a window size of 3 for our final configuration.

\begin{table}[htbp]
\centering
\begin{tabular}{c|ccc}
\hline
\multicolumn{1}{c}{\multirow{2}{*}{$\tau$}} & \multicolumn{3}{c}{MeViS} \\
& $\mathcal{J}$ & $\mathcal{F}$ & $\mathcal{J}\&\mathcal{F}$ \\
\hline
5 & 45.8 & 50.2 & 48.0 \\
10 & 47.3 & 54.3 & 50.8 \\
\textbf{15} & \textbf{49.2} & \textbf{55.6} & \textbf{52.4} \\
20 & 46.9 & 50.5 & 48.7 \\
30 & 43.1 & 48.7 & 45.9 \\
\hline
\end{tabular}
\caption{Ablation study on the keyframe sampling interval ($\tau$) on the MeViS dataset. An interval of $\tau=15$ is selected as our default setting, as it achieves the best performance.}
\label{tab:pai_figre}
\end{table}

\begin{table}[htbp]
\centering
\begin{tabular}{c|ccc}
\hline
\multicolumn{1}{c}{\multirow{2}{*}{Verification Window}} & \multicolumn{3}{c}{MeViS} \\
& $\mathcal{J}$ & $\mathcal{F}$ & $\mathcal{J}\&\mathcal{F}$ \\
\hline
1 & 44.3 & 49.1 & 46.7 \\
\textbf{3} & \textbf{49.2} & \textbf{55.6} & \textbf{52.4} \\
15 & 51.5 & 55.1 & 53.3 \\
\hline
\end{tabular}
\caption{Ablation study on the Verification Window size for the predictive association criterion on the MeViS dataset.}
\label{tab:k_figre}
\end{table}

\subsection{C. More Qualitative Results}
\begin{figure*}[t]
    \centering
    \includegraphics[width=1.0\textwidth]{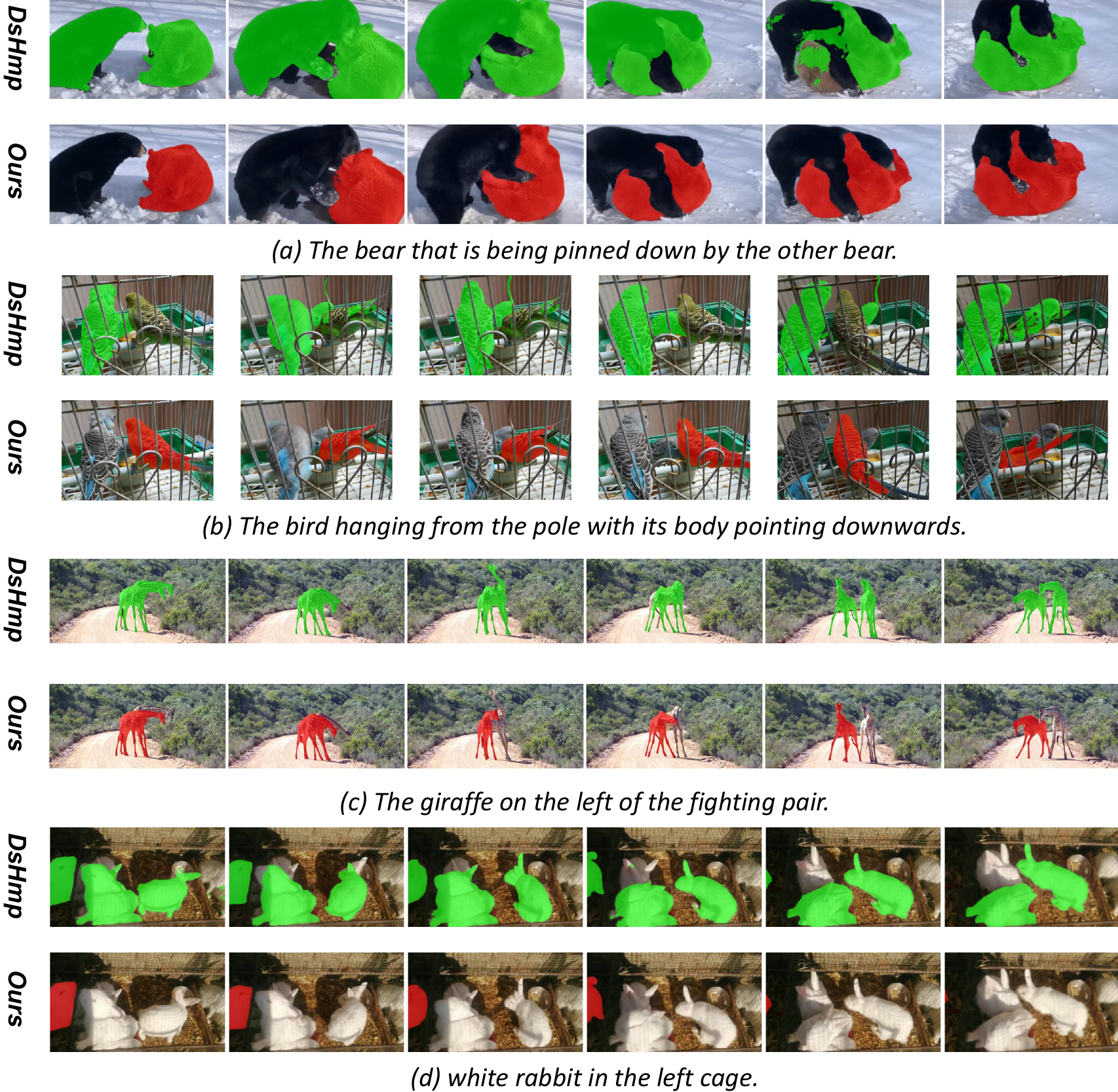}
    \caption{Qualitative comparison between our method and the baseline DsHmp on the MeViS dataset.}
    \label{fig:appendix_fig1}
\end{figure*}
\begin{figure*}[t]
    \centering
    \includegraphics[width=1.0\textwidth]{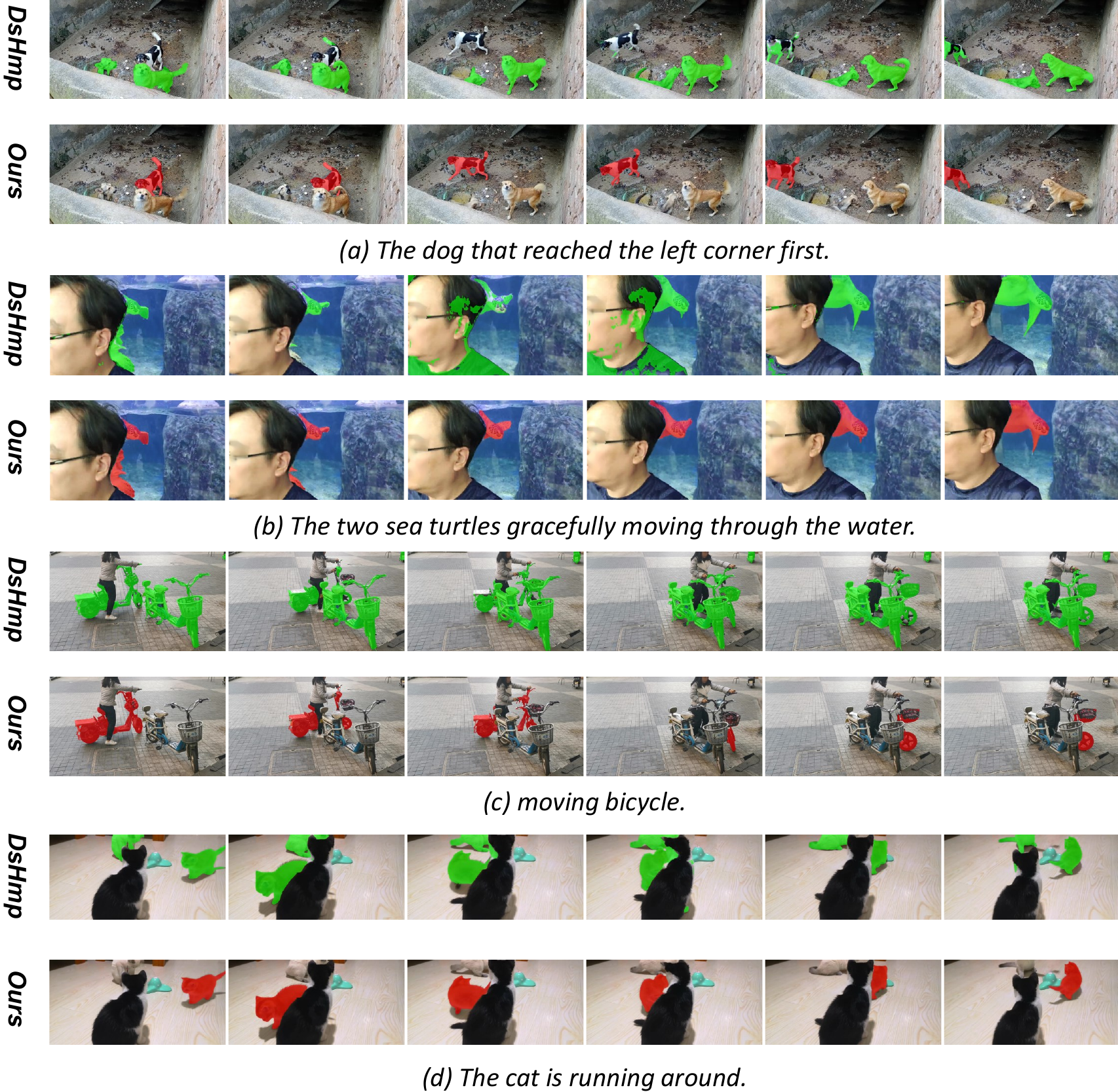}
    \caption{Qualitative comparison between our method and the baseline DsHmp on the MeViS dataset.}
    \label{fig:appendix_fig2}
\end{figure*}

To further demonstrate the robustness and effectiveness of our PARSE-VOS framework, this section provides additional qualitative results on challenging scenarios from the MeViS dataset. These examples visually highlight the superiority of our hierarchical reasoning approach compared to the baseline method, DsHmp\cite{he2024decoupling}.
\end{document}